# CLINICAL DETERIORATION PREDICTION IN BRAZILIAN HOSPITALS BASED ON ARTIFICIAL NEURAL NETWORKS AND TREE DECISION MODELS

Hamed Yazdanpanah[1,2], Augusto C. M. Silva[1], Murilo Guedes[1], Hugo M. P. Morales[1], Leandro dos S. Coelho[3,4], and Fernando G. Moro[1]

[1] *Department of Research, Robô Laura, Curitiba, PR, Brazil*
[2] *Department of Computer Science, University of São Paulo, São Paulo, SP, Brazil*
[3] *Industrial and Systems Engineering Graduate Program, Pontifical Catholic University of Parana, Curitiba, PR, Brazil*
[4] *Electrical Engineering Graduate Program, Federal University of Parana, Curitiba, PR, Brazil*

**ABSTRACT**

Early recognition of clinical deterioration (CD) has vital importance in patients' survival from exacerbation or death. Electronic health records (EHRs) data have been widely employed in Early Warning Scores (EWS) to measure CD risk in hospitalized patients. Recently, EHRs data have been utilized in Machine Learning (ML) models to predict mortality and CD. The ML models have shown superior performance in CD prediction compared to EWS. Since EHRs data are structured and tabular, conventional ML models are generally applied to them, and less effort is put into evaluating the artificial neural network's performance on EHRs data. Thus, in this article, an extremely boosted neural network (XBNet) is used to predict CD, and its performance is compared to eXtreme Gradient Boosting (XGBoost) and random forest (RF) models. For this purpose, 103,105 samples from thirteen Brazilian hospitals are used to generate the models. Moreover, the principal component analysis (PCA) is employed to verify whether it can improve the adopted models' performance. The performance of ML models and Modified Early Warning Score (MEWS), an EWS candidate, are evaluated in CD prediction regarding the accuracy, precision, recall, F1-score, and geometric mean (G-mean) metrics in a 10-fold cross-validation approach. According to the experiments, the XGBoost model obtained the best results in predicting CD among Brazilian hospitals' data.

## 1. INTRODUCTION

Clinical deterioration (CD) is a physiological decompensation, and it happens when a patient undergoes deteriorating conditions or the onset of a severe physiological inconvenience. CD is a leading cause of mortality in hospitals and, in the case of late detection, it can cause organ failure and death (Fleischmann et al., 2016). Conventionally, early warning scores (EWS), such as the Quick Sequential Organ Failure Assessment (qSOFA) (Angus et al., 2016) and the Modified Early Warning Score (MEWS) (Subbe et al., 2006), use only vital signs to determine the CD risk of hospitalized patients.

However, Machine Learning (ML) models are able to utilize laboratory exams, Electronic Health Records (EHRs) data and demographic data in the model to attain more accurate predictions for CD by returning fewer false alarms and more precise detection (Al-Mualemi et al., 2021; Wyk et al., 2019; Wang et al., 2018; Deng et al., 2022). The easier access of EHRs data in hospitals and improved ML strategies encouraged researchers and clinical staff to predict CD in hospitalized patients using automated ML models.

Although there are significant advances in predicting CD by ML models, the majority of studies focused on Intensive Care Unit (ICU) data (Kong et al., 2020; Ibrahim et al., 2020; Moor et al., 2021; Selcuk et al., 2022). Therefore, in this work, ML models are designed to predict CD in patients hospitalized in departments different from ICU. To this end, 103,105 unique attendances out of ICU from thirteen Brazilian hospitals in different states are utilized in the ML models development. The collection period of these data is from March 2015 to July 2021.

In general, when dealing with tabular and structured data, popular ML models outperform artificial neural networks by providing better prediction, higher interpretability, and lower computational cost. In particular, tree-based models, such as Random Forest (RF), Light Gradient Boosting Machine (LightGBM), and eXtreme Gradient Boosting (XGBoost), absorbed more attention among ML models (Yuan et al., 2020; Zabihi et al., 2019; Lyra et al., 2019). Deep neural networks have provided outstanding achievement on unstructured data, such as images, video, audio, and text data (Bengio et al., 2017). However, recently, an extremely boosted neural network (XBNet) has been proposed by combining gradient boosted tree with a feed-forward neural network (Sarkar, 2021). This model uses the feature importance of a gradient boosted tree to update the weights of each layer of the neural network.

For some data sets, it has been shown that this architecture can outperform the traditional ML models (Sarkar, 2021). Therefore, in this work, we compare the performance of the XBNet in CD prediction with some tree-based models, such as XGBoost and RF. Also, to the best of our knowledge, it is the first time the XBNet is applied to the EHRs data for CD prediction. Moreover, the Principal Component Analysis (PCA) is utilized to reduce the data set dimension and to avoid abundant information. Thus, considering the 95% cut-off threshold in the PCA, the number of features decreases from 113 to 73, and the performance of models is compared before and after dimension reduction.

The CD prediction in this work is a binary classification task, where one class is for survival (class 0), and the other one represents death (class 1). Classical metrics for classification tasks, such as accuracy, precision, and recall, are employed to evaluate the performance of models. Furthermore, since our data set is highly class-imbalanced (death rate is almost 4%), some useful metrics for class-imbalanced data set, such as F1-score and geometric mean (G-mean), are reported. Also, all metrics are evaluated in a 10-fold Cross-Validation (CV) framework to prevent overfitting.

The organization of the remainder of this article is as follows. Some related works to the CD prediction are highlighted in Section II. Section III describes a brief description of the employed classifiers and methodology. The experimental results and discussions are presented in Section IV. Finally, conclusions are drawn in Section V.

## 2. RELATED WORKS

Recent advances in the ML field, together with the availability of sources of health and hospital care data, such as EHRs, have generated opportunities for automated medical decision-making to avoid complex problems in health monitoring and to identify clinical deterioration (Ye et al., 2020; Zheng et al., 2017). To this end, several ML models were employed to extract and evaluate different features of EHRs data (Negro-Calduch et al., 2021; Pang et al., 2021).

EHRs data from ICU are utilized in an RF-based model for the early prediction of sepsis (Nakhashi et al., 2019). In (Abromavičius et al., 2019), the patients in ICU are divided into short (less than 9 hours), medium (9 to 60 hours), and long (more than 60 hours) stay in ICU. Based on the patient's stay in ICU, they extracted different vital sign features. Then, gentle adaptive boosting ensemble learning and random under-sampling boosting algorithms are used for sepsis prediction in ICU. Also, in (Oğul et al., 2019), model-based techniques are compared with the instance-based ones by employing elastic time series measures to measure similarity between different instances of vital signs and predict septic shock in ICU.

In (Hu et al., 2019), data are collected from neonatal ICUs and are transformed into images. Then, a convolutional neural network (CNN) is implemented to predict late-onset neonatal sepsis. In (Chang et al., 2019), a recurrent imputation for time series is used to input missing values in vital signs and lab measurements. Then, temporal convolutional neural networks are employed on the imputed data to predict the onset of sepsis. Moreover, in (Wyk et al., 2017), a CNN model is compared to a multilayer perceptron model to detect sepsis, and the CNN model obtained a higher accuracy.

Also, vital signs, laboratory and demographic data are utilized in the early detection of sepsis six hours ahead of time by a bi-directional gated recurrent units model (Wickramaratne et al., 2020). In (Roussel et al., 2019), it is assumed that the prediction of the evolution of vital signs is needed to predict sepsis accurately. Thus, a recurrent artificial neural network is used to predict the vital signs six hours ahead, then the prediction of sepsis in ICU is implemented. Also, in (Demirer et al., 2019), partially observed Markov decision processes are used along with vital signs, laboratory and demographics data to design an artificial intelligence-based sepsis warning system.

## 3. BRIEF DESCRIPTION OF THE CLASSIFIERS

In this section, the XBNet, XGBoost, RF, and PCA are briefly reviewed. These models will be used for the CD prediction in the next section.

### 3.1 XBNet

Recently, XBNet was proposed by combining gradient boosted tree and a feed-forward neural network (Sarkar, 2021). In this algorithm, trees are trained in all neural network layers, and feature importance is obtained by the trees. Then, weight defined by the gradient descent is employed to modify the weights of neural network layers where trees are trained. This approach makes the neural network model robust for tabular data regarding all performance metrics. The adopted optimization strategy in the XBNet is the boosted gradient descent, and it is initialized using the feature importance of gradient boosted trees. Thus, the weights of each layer are updated by the model in the following steps: (i) weights are updated by gradient descent; and (ii) weights are updated by employing the feature importance of gradient boosted trees.

### 3.2 XGBoost

Boosting is an ensemble algorithm that converts a set of weak learners into a strong estimator by sequentially training ML models, in which, for each iteration, the model tries to correct the previous iteration. In gradient boosting, each new iteration is optimized in the residual error of the previous iteration using gradient descent. A popular system for this method is XGBoost, described by (Chen et al., 2016). It introduces a number of novel strategies to optimize learning speeds, enabling it to run ten times faster than gradient boosting machine while maintaining state-of-the-art results.

### 3.3 Random forest

RF classifier is an ensemble learning model that consists of numerous decision trees and was established by (Breiman, 2001). In order to define the split of each node, RF considers only a random sample of the features and calculates the optimal cut-off point for each subset, which results in an ensemble of less correlated decision trees, potentially improving the accuracy of the model. The prediction of an instance is calculated by combining the predictions of all trees (through averaging if the target is numerical or through a majority voting if the target variable is categorical).

### 3.4 PCA

The principal component analysis seeks to express the most significant possible variability of the original features, replacing them with a new smaller set of independent features known as components. These components are linear combinations of original features with the eigenvectors of the variance-covariance matrix of the original features. The eigenvectors point to the direction of more significant variability, which means that a few components are able to retain most of the original variability. More details about principal components analysis can be found in (Jollifa et al., 2016).

## 4. RESULTS ANALYSIS

In this section, the employed data set for generating and evaluating the CD classifiers are described briefly. Then, the results and discussions are presented.

The data set contains 103,105 unique attendances (samples) of patients hospitalized out of ICU. Also, to focus the CD prediction on adults, all patients younger than 18 years old are discarded. The data are collected by EHRs from thirteen Brazilian hospitals from March 2015 to July 2021. The data set contains 77 variables (columns), where 6 of them are categorical variables, such as gender, registered disease, and clinical specialty. Thus, after implementing one-hot encoding, the number of features increases to 113. Regarding vital signs, such as heart rate, temperature, respiratory rate, glucose, oxygen saturation, systolic and diastolic blood pressure, some features describe the last five collections of them. Also, age, days from the last hospitalization, and length of stay are reported in the data set. Moreover, it should be mentioned that the use of this data set is approved by the ethics committee of the corresponding hospitals under protocol number 99706718.9.1001.0098.

All samples in the data set have completed the consultation, i.e., their output is medical discharge or death. Moreover, the designed models are predicting CD events for 12 hours ahead. To this end, the last 12 hours of vital signs before the patient's outcome are removed for each patient. Since the last five collections of vital signs are reported in the data set, and there is a correlation between different collections, some statistical measures of the last five vital signs, such as minimum, maximum, mean, median, and standard deviation (STD), are used as additional features. For patients hospitalized in different hospitals for treatment, information about the time between hospitalizations is utilized. Also, the filling forward imputation technique is adopted to impute missing values using the last collected vital signs. Finally, Table 1 describes the mean, STD, and the missing value percentage of numerical variables of the Brazilian EHRs data set for all samples and different classes.

Table 1. The missing value percentage, mean and STD of numerical variables of Brazilian EHRs data set

| Variables | Total | Survival | Mortality | Missing (%) |
|---|---|---|---|---|
| Heart rate | 79.18 ± 15.06 | 78.60 ± 14.38 | 95.01 ± 22.55 | 11.34 |
| Respiratory rate | 18.11 ± 3.92 | 18.07 ± 3.88 | 19.19 ± 4.91 | 15.16 |
| Diastolic blood pressure | 71.96 ± 11.36 | 72.26 ± 11.10 | 63.76 ± 14.88 | 11.56 |
| Systolic blood pressure | 120.40 ± 18.19 | 120.90 ± 17.78 | 106.64 ± 23.09 | 11.53 |
| Oxygen saturation | 95.87 ± 2.92 | 95.99 ± 2.68 | 92.57 ± 5.84 | 16.12 |
| Capillary blood glucose | 137.66 ± 60.65 | 137.19 ± 59.40 | 145.94 ± 78.84 | 76.37 |
| Temperature | 36.06 ± 0.69 | 36.06 ± 0.68 | 36.20 ± 0.82 | 16.26 |
| Days from entrance | 5.20 ± 9.86 | 4.91 ± 9.04 | 13.47 ± 21.67 | 0 |
| Age | 56.05 ± 18.07 | 55.55 ± 17.93 | 70.55 ± 15.92 | 0 |

Different metrics are used to measure the classifiers' skills in CD prediction. In classification tasks, accuracy, precision, and recall are common metrics to evaluate ML models. Thus, these metrics are reported in this work. However, these metrics are suitable for symmetric data set. The mortality rate is low in our data set (almost 4%), and the data set is highly class-imbalanced. Thus, some appropriate metrics for class-imbalanced data sets such as F1-score and G-mean are reported too.

All metrics are computed in the stratified 10-fold CV approach to avoid overfitting. This approach helps in measuring the ability of ML models with lower bias. The data set is randomly divided into 10 stratified folds of almost the same size to execute CV. Then, the model is trained on 9 folds and is validated on the remaining one fold. This process is repeated until all folds appear one time as a validation set. Finally, the 10-fold CV outcomes are reported via the selected metrics' mean and STD. Furthermore, the hyperparameters of ML models are tuned by the grid search strategy.

In this work, the XBNet model has two hidden layers, where the input and output dimensions of the first layer are 8 and 4, respectively. Also, the input and output dimensions of the second layer are 4 and 2, respectively. For both layers, the bias is set as False. The loss function is the cross-entropy loss, and the optimizer is the adaptive moment estimation (Adam) with the learning rate equal to $3 \times 10^{-4}$. Moreover, the batch size is 32, and the number of epochs is 100.

For the XGBoost model, the number of boosting rounds is 100, and the maximum tree depth for base learners is 6. The subsample ratio of the training instance and the subsample ratio of columns when generating each tree are 0.7 and 0.75, respectively. The $\ell_1$ and $\ell_2$ regularization parameters are adopted as 5. The minimum sum of instance weight required in a child is 9. The minimum loss reduction needed to make a further partition on a leaf node of the tree is 0.3. Finally, the learning rate is chosen as 0.12.

For the RF, the number of trees is 100. The maximum depth of the tree is 18. Also, the minimum number of samples needed to be at a leaf node is 4. Furthermore, MEWS is selected as an EWS candidate for comparison with the ML models since it is conventionally utilized in CD prediction. Also, note that the learning curves in Figs 1, 2, 4, and 5 are generated by taking the average of learning curves between 10-fold CV outcomes.

Fig. 1(a) shows the logistic loss function learning curves for the training and validation sets of the XBNet model during 100 epochs. Moreover, the accuracy learning curves of the XBNet model for the training and validation sets during 100 epochs are shown in Fig. 1(b). Also, the logistic loss function learning curves for the training and validation sets of the XGBoost model are depicted in Fig. 2(a). It can be observed that these two curves are extremely close to each other, and they can attain logistic loss values less than 0.1. Fig. 2(b) presents the accuracy learning curves of the XGBoost model for the training and validation sets. It is worthwhile to mention that besides the visual distance between the training and validation sets learning curves, the overfitting in the model did not happen since the distance between two curves is less than 0.01 (1% of accuracy). In other words, the difference of accuracy between the training and validation sets is less than 1%.

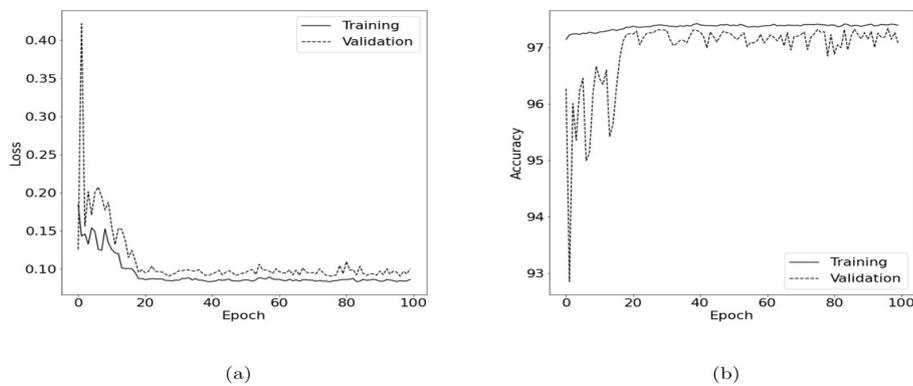

Figure 1. The learning curves of the XBNet model for: (a) logistic loss function versus the number of epochs; (b) accuracy versus the number of epochs.

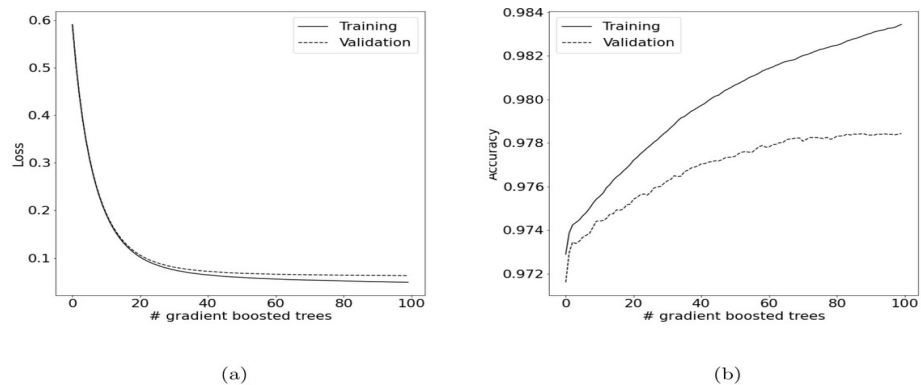

Figure 2. The learning curves of the XGBoost model for: (a) logistic loss function versus the number of gradient boosted trees; (b) accuracy versus the number of gradient boosted trees.

Furthermore, the PCA approach is applied to our data set, which contains 113 features, to reduce the number of features. Fig. 3 shows how many principal components are required to explain the variance in the data set. As presented in this figure, 73 principal components out of 113 ones in the transformed space are required to explain 95% of variance in the data set. It shows that this data set contains informative variables, and most features provide unique information about the data set. Therefore, the first 73 principal components are used as the new data set for training and validating the XBNet and XGBoost models.

The logistic loss function learning curves of the XBNet model for the training and validation sets when using 73 principal components are depicted in Fig. 4(a). Also, Fig. 4(b) presents the accuracy learning curves of the XBNet model for the training and validation sets when 73 principal components are utilized. As can be observed in Figs. 1 and 4, the XBNet model converges faster when it is applied to 73 principal components rather than 113 features. Also, the logistic loss function learning curves of the XGBoost model for the training and validation sets when employing 73 principal components are shown in Fig. 5(a). In addition, when 73 principal components are used as inputs, the accuracy learning curves of the XGBoost model for the training and validation sets are presented in Fig. 5(b). By comparing Figs. 2 and 5, it is evident than the distance between the training and validation curves is higher in Fig. 5 than that of in Fig. 2. Therefore, overfitting is more likely when using the XGBoost on the principal components of the employed data set.

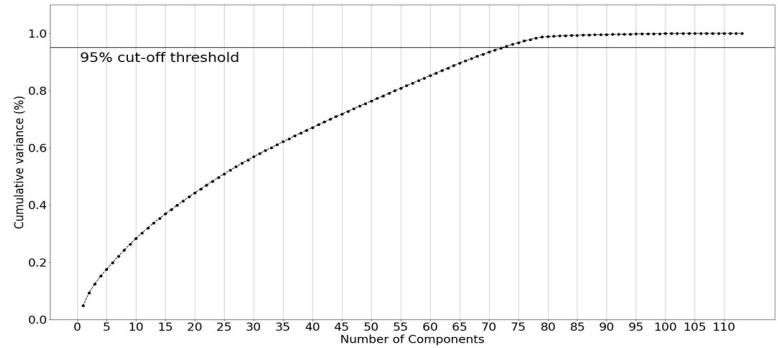

Figure 3. 73 of 113 principal components are required to explain 95% of variance in the data set.

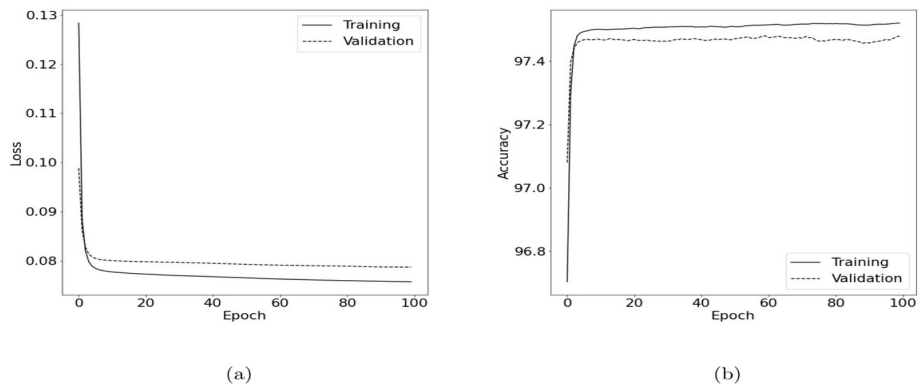

Figure 4. The learning curves of the XBNet model, when the number of features is reduced by the PCA, for: (a) logistic loss function versus the number of epochs; (b) accuracy versus the number of epochs.

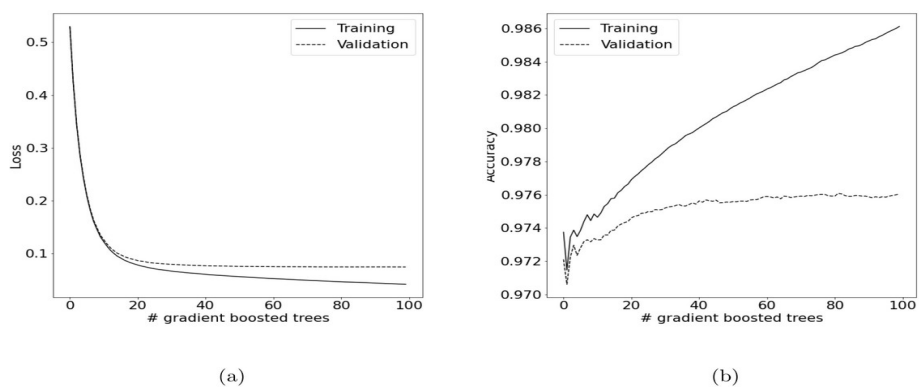

Figure 5. The learning curves of the XGBoost model, when the number of features is reduced by the PCA, for: (a) logistic loss function versus the number of gradient boosted trees; (b) accuracy versus the number of gradient boosted trees.

Moreover, as another tree-based model candidate for predicting CD, the RF is applied to the data set before and after executing the PCA. For the XBNet, XGBoost, and RF models before and after applying

PCA, and MEWS the mean and STD of accuracy, precision, recall, F1-score, and G-mean are reported in Table 2. As presented in this table, when the XGBoost model is applied to the original data set (without using principal components), it outperforms other tested models according to all the evaluated metrics, except the precision. In CD applications, having higher recall is more valuable than obtaining higher precision, and the XGBoost model can bring the highest recall among the tested models. Also, since our data set is highly class-imbalanced, the G-mean and F1-score values are of utmost importance to measure the superiority of models in CD prediction. Furthermore, it can be observed that the values of metrics for MEWS are significantly lower than those for the ML models, except the mean of G-mean for the RF+PCA.

Table 2. Accuracy, precision, recall, F1-score, and G-mean of algorithms for 10-fold CV

| Algorithms | Accuracy | | Precision | | Recall | | F1-score | | G-mean | |
|---|---|---|---|---|---|---|---|---|---|---|
| | Mean | STD | Mean | STD | Mean | STD | Mean | STD | Mean | STD |
| XBNet | 0.971 | 0.0057 | 0.635 | 0.1154 | 0.404 | 0.0381 | 0.485 | 0.0390 | 0.632 | 0.0289 |
| XBNet + PCA | 0.975 | 0.0009 | 0.755 | 0.0195 | 0.376 | 0.0380 | 0.501 | 0.0335 | 0.611 | 0.0308 |
| XGBoost | **0.978** | 0.0011 | 0.800 | 0.0229 | **0.482** | 0.0247 | **0.601** | 0.0237 | **0.692** | 0.0177 |
| XGBoost + PCA | 0.976 | 0.0011 | 0.774 | 0.0293 | 0.409 | 0.0289 | 0.535 | 0.0288 | 0.638 | 0.0227 |
| RF | 0.976 | 0.0011 | **0.890** | 0.0302 | 0.326 | 0.0289 | 0.476 | 0.0327 | 0.570 | 0.0250 |
| RF + PCA | 0.973 | 0.0006 | 0.874 | 0.0250 | 0.239 | 0.0160 | 0.375 | 0.0207 | 0.488 | 0.0166 |
| MEWS | 0.928 | 0.0023 | 0.187 | 0.0105 | 0.338 | 0.0192 | 0.241 | 0.0125 | 0.566 | 0.0159 |

The required execution time for the XBNet, XGBoost, and RF models are 23724, 18, and 29 seconds, respectively. As can be observed, the XGBoost model has the fastest execution time (18 seconds), and the RF requires a few more seconds to be executed. However, the execution time of the XBNet model is much higher than other models. Indeed, it needs more than six hours to be implemented.

## 5. CONCLUSION

In this paper, EHRs and demographic data of more than 100,000 samples of hospitalized patients in Brazilian hospitals have been utilized for CD prediction. All patients were hospitalized in departments different from ICU. The XBNet model, as a neural network candidate, has been employed for predicting CD using tabular data, and its performance has been compared to the XGBoost and RF models as two tree-based models. Also, the PCA approach has been adopted to reduce the number of features, and the XBNet, XGBoost, and RF have been trained on the transformed data set to verify whether the PCA technique can improve the performance of models. The XGBoost obtained the best results among all tested models by resulting in the highest accuracy, recall, F1-score, and G-mean. Furthermore, the XGBoost model had the minimum execution time compared to the XBNet and RF models, and the XBNet model needed much more time to be executed.

In this work, it has been shown that the XBNet model cannot hit the XGBoost model performance on Brazilian hospitals' data set. However, in future research directions, more effort will be put into the feature engineering task to verify if, with new and more creative features, the XBNet model can obtain a superior performance to conventional ML models.